\documentclass[10pt,twocolumn,letterpaper]{article}

\usepackage{iccv}
\usepackage{times}
\usepackage{epsfig}
\usepackage{graphicx}
\usepackage{amsmath}
\usepackage{amssymb}

\usepackage[inline]{enumitem}
\usepackage{subfigure} 
\usepackage{multirow}
\usepackage{xcolor}
\usepackage{colortbl}
\usepackage{float}
\newcommand{\colorline}[2]{\arrayrulecolor{#2}\cline{#1}\arrayrulecolor{black}}
\newcommand{\ccb}[1]{\cellcolor{cyan!#1}}
\newcommand{\ccr}[1]{\cellcolor{red!#1}}
\newcommand{\ccg}[1]{\cellcolor{green!#1}}
\def\modelname{HAVSampler} 
\usepackage[pagebackref=true,breaklinks=true,letterpaper=true,colorlinks,bookmarks=false]{hyperref}

\iccvfinalcopy % *** Uncomment this line for the final submission

\def\iccvPaperID{11318} % *** Enter the ICCV Paper ID here

\ificcvfinal\fi

\begin{document} 

%%%%%%%%% TITLE
% \title{Rethinking Point Sampling\\ in Large-Scale Point cloud for Real-time Application}
\title{Hierarchical Adaptive Voxel-guided Sampling\\ for Real-time Applications in Large-scale Point Clouds}

% \author{Junyuan Ouyang 
% \\Institution1\\
% \and
\author{Junyuan Ouyang \quad Xiao Liu \quad Haoyao Chen\\
Harbin Institute of Technology, Shenzhen\\
{\tt\small 21s053086@stu.hit.edu.cn}
}
\maketitle
\ificcvfinal\thispagestyle{empty}\fi
%%%%%%%%% ABSTRACT
\begin{abstract}
% 第三句和实验末尾的breaking bottlenect可以删去
While point-based neural architectures have demonstrated their efficacy, the time-consuming sampler currently prevents them from performing real-time reasoning on scene-level point clouds.
Existing methods attempt to overcome this issue by using random sampling strategy instead of the commonly-adopted farthest point sampling~(FPS), but at the expense of lower performance.
So the effectiveness/efficiency trade-off remains under-explored.
In this paper, we reveal the key to high-quality sampling is ensuring an even spacing between points in the subset, which can be naturally obtained through a grid.
Based on this insight, we propose a hierarchical adaptive voxel-guided point sampler with linear complexity and high parallelization for real-time applications.
Extensive experiments on large-scale point cloud detection and segmentation tasks demonstrate that our method achieves competitive performance with the most powerful FPS, at an amazing speed that is more than 100 times faster.
This breakthrough in efficiency addresses the bottleneck of the sampling step when handling scene-level point clouds. Furthermore, our sampler can be easily integrated into existing models and achieves a 20$\sim$80\% reduction in runtime with minimal effort. The code will be available at \small{\url{https://github.com/OuyangJunyuan/pointcloud-3d-detector-tensorrt}}

\end{abstract}

% While point-based neural architectures have demonstrated their efficacy in dealing with unordered and non-structured data, they have few scalability due to adopt time-consuming point samplers. 

% Existing methods attempt to address this problem at the expense of lower performance, replacing the widely used farthest point sampling (FPS) that has secondary complexity with an efficient random sampling strategy. 
% Existing methods attempt to address this problem via replacing the mainstream time-consuming Farthest Point Sampling (FPS) with an efficient random sampling strategy, at the expense of lower performance. 

% Thus, how to balance the performance and efficiency is still under explored. 
% In this work, we reveal the essence of high-quality sampling is the even distance between points in subset, which can be obtained naturally by grid. 

% Based on this observation, , we propose a hierarchical adaptive voxel-guided point sampler with linear complexity and GPU-friendly. 

% Extensive experiments on large-scale point cloud detection and segmentation tasks demonstrate both the efficiency and effectiveness of our approach. 

% Our approach is so fast to breaks the sampling bottleneck of point-based methods when applied large-scale point clouds.

%%%%%%%%% BODY TEXT
\section{Introduction} 
Point cloud learning has attracted increasing attention since it benefits 3D scene understanding.
It can be divided into grid-based methods and point-based methods based on the different representations.
This work aims to alleviate the applicability limitations of point-based neural architectures by presenting a lightweight yet powerful point sampler. 
% recent works mainly focus on network designation, while sampling step blocks point-based method to be used in large-scale point cloud.
% recent works mainly focus on network designation, while sampling step blocks the capility of point-based method due to unignoreable comsumption when handling large-scale point cloud.
% 现有工作关注与网络设计，然而采样步骤限制了点方法在大规模点云上的应用，因为此时不可忽略的耗时占据大部分推理时间。

Sampling is a critical operation in most existing point-based networks. 
Existing samplers can be summarized into learning-based methods \cite{zhang2022not,liu2021group,lang2020samplenet,yang2019modeling,dovrat2019learning,cheng2022meta,li2022psnet} and learning-free methods \cite{qi2017pointnet,xu2020grid,nezhadarya2020adaptive,li2022adjustable,yang2022graph}.
It keeps a subset with fewer surviving points as the network goes deeper and wider to reduce memory and computation. 
Since a subset of preceding layer is fed to successor layer by layer, sampling has a significant impact on the final task performance. 
So how to extract a representative subset from raw point clouds while maintaining high efficiency is a key problem, especially for real-time applications of large-scale point clouds.

\begin{figure}[!t]
\centering
\includegraphics[width=1.0\linewidth]{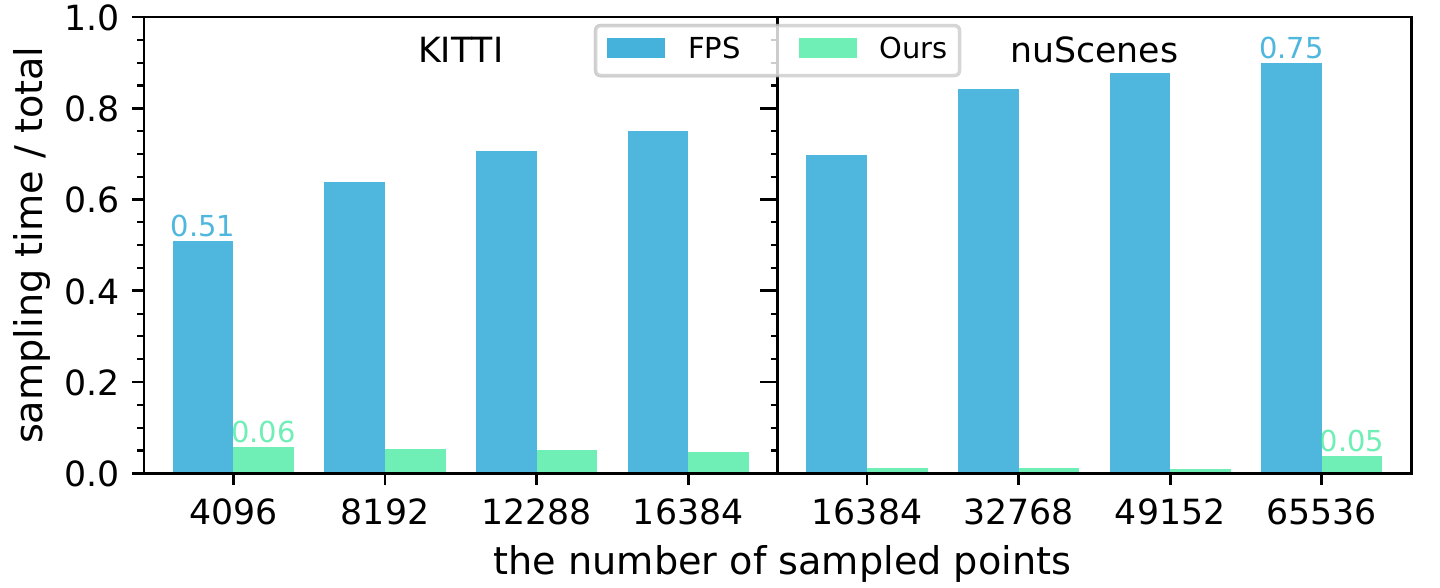}
\caption{Statistical results of sampling latency on the KITTI and nuScenes datasets for detection tasks. Blue bars indicate that the sampler dominates most of the runtime. Green bars show that our approach can significantly break this efficiency bottleneck.}
\label{fig1}
\end{figure} 
% Although these methods have made remarkable progress, 
Although remarkable progress has been made in sampling, it is hard to balance the efficiency and effectiveness when it comes to large-scale point clouds, especially for the first downsampling layer. 
For the sake of higher performance, FPS is the primary method used in many point cloud tasks such as detection \cite{shi2019pointrcnn} and segmentation \cite{qi2017pointnet}. 
However, it takes up to half the inference time when dealing with $10^5$ points as shown in Figure \ref{fig1}. 
And for $10^6$ points, the output of a 32-beam LiDAR is infeasible due to its quadratic complexity and iterative algorithm. 
Therefore, low-latency random point sampling (RPS) is adopted in scenarios where efficiency is critical but suffers from low sampling quality. 
The learning-based approaches cannot be used in the first downsampling layer due to the lack of sufficient features and the huge sampling probability tensor. 

To tackle this problem, we first explore how FPS can benefit the task network to achieve better final performance and then design a more economical algorithm to approximate it.  
By comparing various algorithms, we discover that the even spacing between points in the subset is the key to the success of FPS, rather than the uniform density of points presented in previous work\cite{qi2017pointnet,qi2017pointnet++}, as inverse density sampling~\cite{ning2021density} (IDS) also results in poor performance.
The uniform distance protects the points in the subset from scattering and clustering with each other, making it less messy~(see~Section~\ref{sec4-6}). 
This feature allows the geometric information of the original scene to be preserved as much as possible, which is important for downstream detection and segmentation tasks that more focus on shapes. 
It can be imagined that if we can obtain this property in an efficient way, the performance should be close to FPS. And fortunately, the grid naturally satisfies that because the distance between cells in the grid is constant. 

Given the observation mentioned above, we design an adaptive voxel-guided point sampler that hierarchically selects points closest to their grid centers. 
Our approach has linear time complexity, can be run in parallel, and is both permutation invariant and deterministic. 
In summary, our main contributions are as follows: 

% Based on the above observation, we design an adaptive voxel-guided point sampler to hierarchically select points that are closest to their grid centres. 
% Our strategy with linear complexity can run in parallel, and is permutation invariant and deterministic. 
% We summarize our main contributions as follows:
\begin{itemize}\vspace{-0.05in}
\item We revisit several existing sampling techniques and reveal that the key to achieving high-quality sampling is to obtain a uniformly distributed subset. \vspace{-0.05in}
% \item We revisit several sampler and reveal the key to high-quality sampling is a uniformly distributed subset. \vspace{-0.05in}

\item  Our proposed \modelname{} has a linear time complexity and is parallelizable, making it suitable for real-time applications. Additionally, it is easy to integrate into existing models, making them run up to 5$\times$ faster.

% \item  Our proposed \modelname{} has a linear time complexity and is parallelizable, making it suitable for real-time applications. Additionally, it is easy to integrate into existing models, thus extending their scalability.
% \item The proposed \modelname{} is linear complexity and parallelizable for real-time applications. It is easy to integrate to extend the scalability of existing models. 
\vspace{-0.05in}

\item Our experimental results on various datasets and tasks demonstrate that our method is comparable to or even better than the state-of-the-art method FPS at~$10^2\sim10^4\times$ faster. This breaks the efficiency bottleneck when using point-based models in large point clouds.
% Our experimental results on various datasets and tasks show that our method achieves comparable or even better performance than the state-of-the-art method FPS at 10-20 times faster. This breaks the efficiency bottleneck when using point-based models in large point clouds.
% \item Our experimental results on various datasets and tasks demonstrate that our method achieves comparable or even better performance than FPS at the speed of RPS. This breaks the efficiency bottleneck when using point-based networks in large point clouds.
% \item Experimental results on various datasets and tasks manifest that our strategy achieves comparable or even better performance than FPS at the speed of RPS, breaking the efficiency bottleneck when using point-based networks in large point clouds.
\end{itemize}

\section{Related Work}
% We propose a sampler for alleviating the efficiency problem of the existing point-based 3D learning architectures. 
\paragraph{Point-based Neural Architectures } aim to learn permutation invariant features from unordered point clouds. 
Serving as feature extractors, point-based models are widely adopted in many downstream tasks, i.e., point cloud detection~\cite{chen2022sasa,shi2019pointrcnn,yang20203dssd,zhang2022not}, motion segmentation~\cite{sun2020pointmoseg}, tracking~\cite{zheng2022beyond,wang2020pointtracknet} and registration~\cite{aoki2019pointnetlk}. 
The pioneering work PointNet~\cite{qi2017pointnet} proposes using pointwise MLPs and max pooling to extract robust global features. Succeeding work PointNet++~\cite{qi2017pointnet++} designs a canonical paradigm that hierarchically samples subsets, groups neighboring points and aggregates contexts to obtain local features. 
Due to the success of this paradigm, most of the following works keep this process, such as pointwise models~\cite{qian2022pointnext,ma2022rethinking}, graph-based method~\cite{xu2020grid,landrieu2018large,Zhao_2019_CVPR,wang2019dynamic}, point convolution~\cite{Wu_2019_CVPR,li2018pointcnn,Liu_2019_CVPR} and recent transformer-based methods~\cite{engel2021point,zhao2021point}. 
Obviously, as a first step, sampling has a significant impact on the performance of following feature extraction and task-specific network. 
This work benefits point cloud applications by presenting an effective and efficient sampling strategy.
% A handful of works including 
\paragraph{Learning-free Point Cloud Samplers} can be used for filtering, pre-processing or as a basic component of point-based networks. 
RPS~\cite{hu2020randla,qiu2021semantic,shuai2021backward} is mainly adopted in large-scale point cloud or real-time applications attributed to its high efficiency.
IDS~\cite{groh2019flex,yang2022graph} tends to probabilistically select the points with lower densities in order to preserve sparse points. 
Normal space sampling~\cite{diez2012hierarchical} can preserve the edge of point clouds well. 
Random voxel sampling (RVS) divides points into voxels with a fixed grid and preserves the centroid of the voxels. 
As one of the most advanced samplers, FPS ~\cite{qi2017pointnet,qi2017pointnet++} is the first choice for many state-of-the-art models. 
It iteratively selects the point furthest away from the sampled point and updates subset, providing more complete scene coverage. 
Subsequent works enhance FPS by considering features~\cite{yang20203dssd}, density~\cite{ning2021density,liu2021group} and semantic~\cite{chen2022sasa}\looseness=-1. 

Although great progress has been made in this area, the trade-off between latency and performance is still a challenge. % especially for large point cloud. 
For object-level point clouds, FPS is the best choice to minimize performance degradation caused by sampling. 
When confronted with scene-level points, its latency is non-negligible and even dominant, making it unacceptable. 
Some works~\cite{li2022adjustable,zhang2022not} attempt to alleviate this computationally inefficient problem by sorting points or dividing scenes into partitions. 
However, they do not change the mechanics of the sequential FPS and still cannot be executed in parallel.
While employing RPS~\cite{hu2020randla} can significantly reduce time consumption, it has a poor performance. 
Because RPS misses most of the points in the sparse region, it does not capture enough information. 
IDS~\cite{ning2021density,yang2022graph} can preserve sparse points but conducts time-consuming k-nearest neighbor (KNN) to estimate density. 
Efficient voxel sampling can smooth uneven density, but the performance is also not ideal.
This work solves this difficult balance that hinders the use of point-based networks in many applications. 

\paragraph{Learnable Point Cloud Samplers.}
Recent research has explored how to adaptively sample the points that are critical for downstream tasks in a learnable way to improve performance. 
Some works elaborately enhance learning-free sampler with learnable weights~\cite{zhang2022not,liu2021group,nezhadarya2020adaptive} or task-oriented loss~\cite{cheng2022pcb}. % having less generality. 
Some methods~\cite{dovrat2019learning,lang2020samplenet} generate subsets directly from MLPs.
Others~\cite{yang2019modeling,li2022psnet} predict a probability matrix with shape of target number $\times$ source number, and make sampling step differentiable by Gumbel-Softmax trick. 

However, learnable fashion is not suitable for us. 
Careful initialisation~\cite{li2022psnet} of these samplers is required, otherwise all points are invisible to subsequent networks at the early training. 
And the huge probability matrix will occupy all memories when intake large-scale point cloud. 
More importantly, we want to address the latency/performance trade-off, especially for first-layer sampling in large point clouds.
But we lack sufficient features to sample from the source because we sample exactly for extracting finer features. 

\begin{figure*}[!t]
\centering
\includegraphics[width=0.90\linewidth]{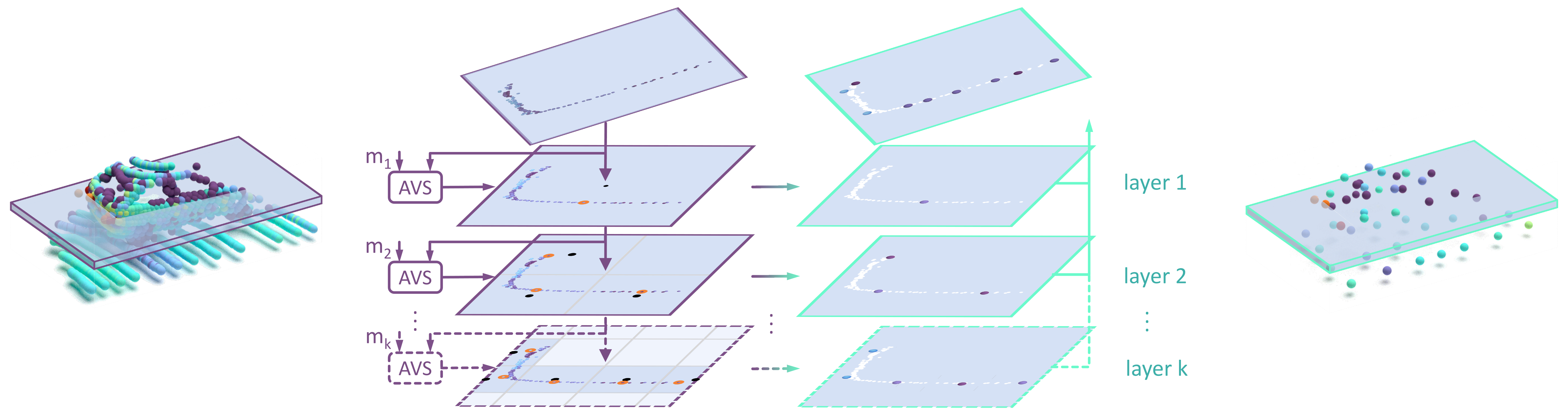}
\caption{Illustration of the overall architecture. \modelname{} contains k layers with different voxel sizes, each consisting of an adaptive voxel searching (AVS) module (see Section~\ref{sec3-2}) and center-closest selection strategy (see Section~\ref{sec3-3}).
The leftmost is the original point cloud of a car. 
The point cloud on the rightmost side is the result of sampling.
We show a 2D slice of a car for easier understanding. 
}
\label{fig2}
\end{figure*}
\section{HAVSampler}
We first formulate point sampler and revisit previous work to explore the key for sampling. Next, we describe the overall design of \modelname{}, including the center-closest sampling strategy and adaptive voxel searching module.

\subsection{Problem Formulation and Revisiting Study\label{sec3-1}}
Point sampling is an essential component of point cloud reasoning, combined with grouping and aggregation to form a basic block. 
It progressively downsamples source points to build deeper networks and acts as a group center to aggregate local features. 
For a given set of source points $\mathcal{P}=\{\mathbf{p}_i \in \mathbb{R}^{3}\}_{i=1}^n $, where each point consists of three coordinates, the point sampler $S$ is expected to extract a subset $\mathcal{P}_s=\{\mathbf{p}_j\in \mathbb{R}^{3}\}_{j=1}^m$ from $\mathcal{P}$.
The low-quality sampled result will damage the following feature extractor and block all the other downstream task heads. 
Therefore, we aim to design an efficient yet powerful sampler for the real-time application of large-scale point clouds.

We revisit the superior FPS and make an in-depth qualitative analysis to find what property makes it stronger. 
%On this basis, we design a more efficient strategy to approximate this property. 
FPS divides the source set $\mathcal{P}$ into two subsets, unselected points $\mathcal{P}_{us}$ and selected points $\mathcal{P}_s$. 
In the beginning, a point is selected at random from the source set to initialize the selected set, and the rest of the source points are considered to be the unselected part. 
Next step, the point-to-set distance 
$d_k=\min{\left\{ \lVert \mathbf{p}_k-\mathbf{p}_j \rVert \mid \mathbf{p}_j\in \mathcal{P}_s \right\}}$ 
is calculated for each unselected point $\mathbf{p}_k \in \mathcal{P}_{us}$ to the selected set $\mathcal{P}_s$, and the point with the furthest distance is transferred from the unselected set to the selected set. The above procedure is repeated until the selected subset contains enough $m$ points. Let $\mathcal{P}_{us}^{(i)}$ denote the unselected set at i-th iteration, then $\mathcal{P}_s$ can be formulated as 
\begin{equation} \label{eq:fpsupdate}
\begin{aligned}
\mathcal{P}_s&=\left\{
\mathop{\textrm{argmax}}\limits_{\mathbf{p_k}} \left\{d_k \mid \mathbf{p}_k\in\mathcal{P}_{us}^{(i)} \right\}
\right\}_{i=1}^m.
\end{aligned}
\end{equation}

\begin{figure}[!t]
\centering

\includegraphics[width=1.0\linewidth]{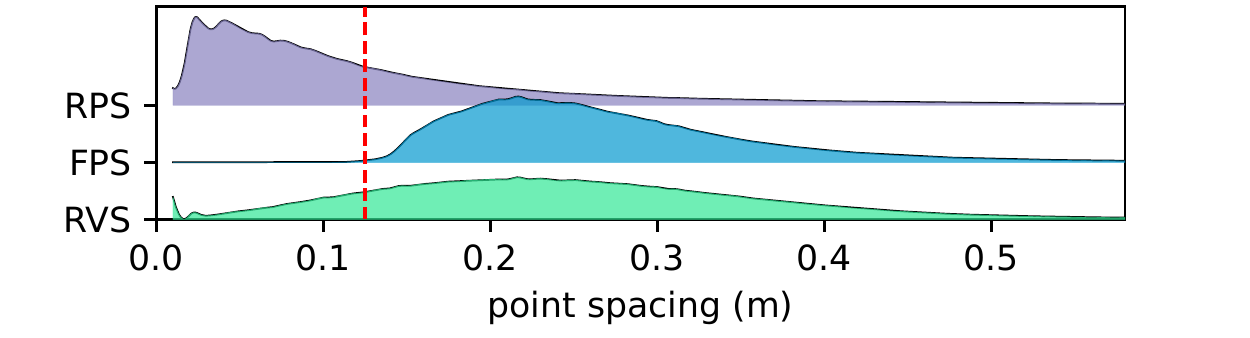}
    \caption{The distance distributions of points in different sampled subsets. The red line marks the minimum spacing of FPS results.}
\label{fig3}
\end{figure}

\bigskip
Since we select the furthest points in $\mathcal{P}$ to form $\mathcal{P}_s$ by Eq.\ref{eq:fpsupdate}, the distances among points in the subset $\mathcal{P}_s$ have a lower bound (see supplementary materials for proof), which can be seen in Figure~\ref{fig3}. 
Besides, the point spacing of FPS results has a higher mean value. 
Larger mean spacing means more scattered sampling results and wider coverage, which maintains a more complete geometry of the raw scene. 
This complete geometry is critical for the performance of detection and segmentation. 
And RVS has a similar distribution to FPS, except for a smaller lower bound. 
In contrast, the spacing distribution of the points in RPS results mostly falls in smaller values, making sampled points cluster together and breaking the raw scene.

Overall, more scattered sampling leads to higher performance. 
If a similar spacing distribution of points in the subset could be achieved in a more efficient way, we can guess that it would have the same high performance as FPS. 
Fortunately, the spacing between the grids is naturally constant. 
As shown in Figure~\ref{fig3}, grid-based RVS has a more even spacing than RPS. 

\subsection{Voxel-guided Sampling\label{sec3-2}}
Inspired by the above analysis, we design an efficient voxel-guided sampler with even point spacing and high quality. 
Since point clouds scanned in 3D space are very sparse\, with typically only 0.1\% of the voxels containing points \cite{Mao_2021_ICCV}, we use a sparse approach to voxelize the point cloud to reduce overhead. 
After that, we select one point from each non-empty voxel to form the final result. 
\begin{figure}\centering   
\subfigure[average] {
 \label{fig4-1}     
\includegraphics[width=0.22\linewidth]{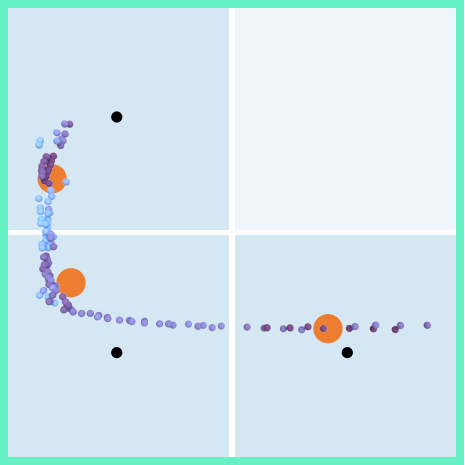}  
}    
\subfigure[random] { 
\label{fig4-2}     
\includegraphics[width=0.22\linewidth]{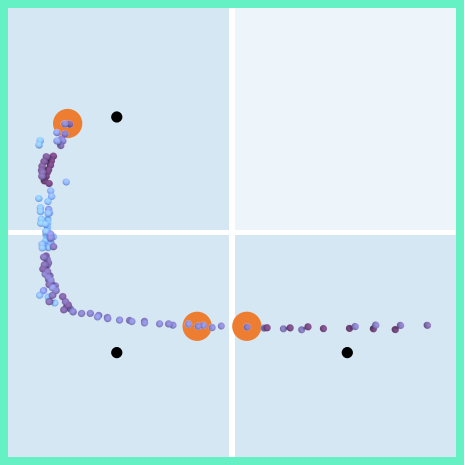}   
}
\subfigure[centre] { 
\label{fig4-3}     
\includegraphics[width=0.22\linewidth]{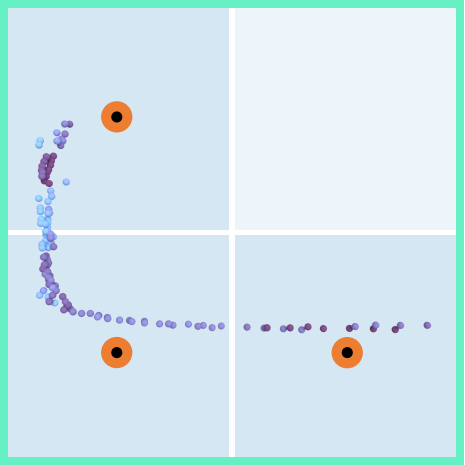}   
}
\subfigure[center-closest] { 
\label{fig4-4}     
\includegraphics[width=0.22\linewidth]{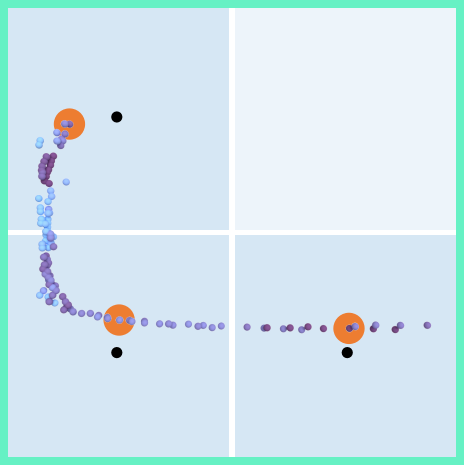}   
}
\caption{Comparison of different point selections. Deep and light blue grids indicate non-empty and empty voxels, respectively. Black dots is grid center. Orange circles highlights the selections. }
%Comparison of different sampling strategy. Deep blue and light blue grids indicate non-empty and empty voxels, respectively. Small black dots is grid center. Orange circles highlights the selections. Our approch have a better sampling quality.
\label{fig4}
\end{figure}
\paragraph{Sparse Voxelization via Voxel-Hashing.} 
Formally, for a given input point cloud $\mathcal{P}$, expected sampling number $m$ and voxel size $\mathbf{v}=\left(v_x,v_y,v_z\right)$, we compute the voxel coordinate $\mathbf{g}_i$ for each $\mathbf{p}_i=\left(x_i,y_i,z_i\right)\in\mathcal{P}$ as shown in
\begin{equation} \label{eq:voxelcoords}
\begin{aligned}
\mathbf{g}_i
=\left(u,v,w\right)
=
\left(
\lfloor\frac{x_i}{v_x}\rfloor,
\lfloor\frac{y_i}{v_y}\rfloor,
\lfloor\frac{z_i}{v_z}\rfloor
\right). 
\end{aligned}
\end{equation}
After that, we need to find the all non-empty voxels to perform sampling. 
Some works \cite{openpcdet2020} filter out the duplicate elements in $\{\mathbf{g}_i\}$ by the $\textrm{unique}$ operation that with a computational complexity of at least $\mathcal{O}(n\log{n})$. 
Instead of unique operation, we use hashtable $\mathcal{T}$ to implement the integer coordinates de-duplication with linear complexity. 
For each~$\mathbf{p}_i$, it fills the table at slot $c_i=\textrm{hash}\left(\mathbf{g}_i\right)$. 
Finally we obtain the all non-empty voxels by traversing the $\mathcal{T}$.
Note that the above steps can be executed in parallel. 

\paragraph{Center-closest Point Sampling.} 
One problem remains to be solved here: which point in a non-empty voxel should we choose to represent that voxel? 
Previous works mainly select the average, center, or random points of a voxel, as shown in Figure~\ref{fig4}. 
We observe three drawbacks to these: 
\begin{enumerate*}[label=(\arabic*)]
\item the average value or the center of the voxel is a fake point that does not belong to the original set;
\item random selection or average points will reduce the spacing between points;
\item the use of centers introduces quantization errors. 
\end{enumerate*}

Taking these three aspects into account, we propose selecting the point closest to the voxel center as the representative. 
Formally, we extend $\mathcal{T}$ to store pairs of point index and the distance $(i,d_i)$. 
For each $\mathbf{p}_i\in\mathcal{P}$, its distance to corresponding voxel centre is computed by $d_i=\lVert\mathbf{p}_i-\mathbf{v}\odot\left(\mathbf{g}_i+0.5\right)\rVert$, where $\odot$ is element-wise multiplication. 
To finger out the closest points, we update $\mathcal{T}(c_i)$ with $(i,d_i)$ pair if $d_i$ is smaller than the existing distance value at slot $c_i$, which means $\mathbf{p}_i$ is closer to the center than other points in the same voxel. 
As illustrated in Figure~\ref{fig4}, the points sampled by our strategy are included in the source set and have no quantification errors. 
Besides, by being closest to the grid center, the spacing between sampled points is scattered as much as possible. 
Therefore, our strategy can approximate the point distribution of FPS in an efficient manner and achieves high performance. 

\begin{figure}[!t] \centering   
%%%%%%%%%%%%%%%%%%%%%%%%%%%%%%%%%%%%%%%%%%%%%%%%%%%%%%%%%%%%%%%
\subfigure[] {
 \label{fig5-1}     
\includegraphics[width=0.45\linewidth]{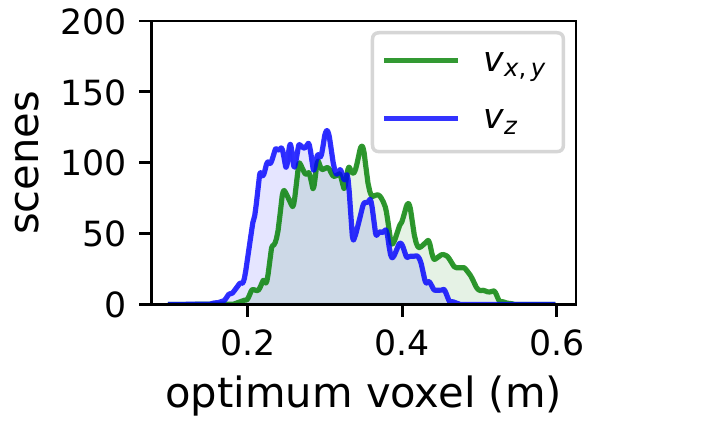}  
}    
\subfigure[] { 
\label{fig5-2}     
\includegraphics[width=0.45\linewidth]{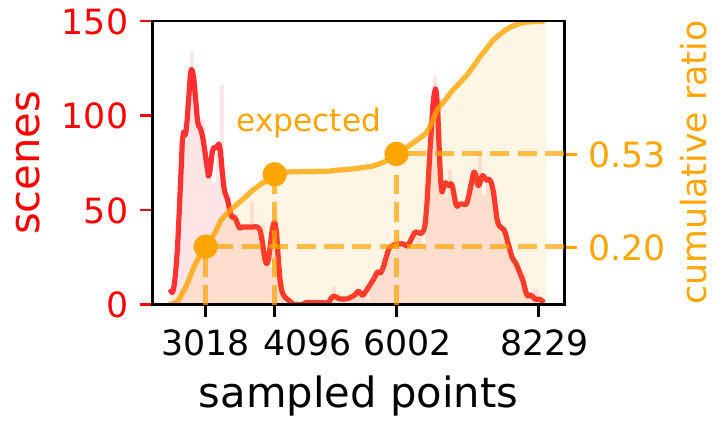}   
}
% \subfigure[] { 
% \label{fig5-3}     
% \includegraphics[width=0.21\linewidth]{iccv2023AuthorKit/img/voxel_and_npts1.pdf}     
% }
%%%%%%%%%%%%%%%%%%%%%%%%%%%%%%%%%%%%%%%%%%%%%%%%%%%%%%%%%%%%%%%

\caption{
Statistical results on KITTI dataset. 
(a) Distribution of optimum voxel size when sampling 4096 points. 
(b) Red curve shows the distribution of the number of non-empty voxels, i.e., the number of sampled points, when fixing the voxel size to $[0.34,0.34,0.30]$ m, mean of the optimum voxel in (a); 
The orange curve is the normalized area under the red curve. 
}     
\label{fig5} 
\end{figure}
\subsection{Adaptive Voxel Initialization Strategy\label{sec3-3}}
Although we can extract a subset with a scattered point distribution, there is still a hyper-parameter, the voxel size $\mathbf{v}$, that needs to be tuned. 
Most previous works \cite{yan2018second,lang2019pointpillars,yin2021center} use a fixed voxel size across different scenes. 
However, the scale of different point clouds can vary greatly, resulting in too many or too few non-empty voxels and sampled points. 
As shown in Figure~\ref{fig5-1}, the most suitable voxel size, i.e., the number of non-empty voxels matches the expected sampling number, can widely range between 0.2$\sim$0.5 m for different scenes. 
Sampling with the mean of the optimal voxel, Figure~\ref{fig5-2} shows that there will be about 20\% scenes with 25\% undersampling rate and up to 50\% scenes with 50\% oversampling rate. 
Therefore, the sampled points need to be resampled or cut off, resulting in subsets containing redundant points or missing information. 
% Although we can extract a subset with a sufficiently scattered point distribution by the above strategy. 
% it can be noticed that in our sampler there is still a hyper-parameter, the voxel size $\mathbf{v}$, that needs to be tuned. 
% Most previous works \cite{yan2018second,deng2021voxel,zhou2018voxelnet,lang2019pointpillars,yin2021center} use a fixed voxel size across different scenes. 
% However, the scale of different point clouds can vary significantly. 
% Thus, fixed voxels may produce too many or too few non-empty voxels. 
% As shown in Figure~\ref{fig5-1}, the most suitable voxel size for each scene in KITTI datasets~\cite{geiger2012we} can widely distribute between 0.2 m and 0.5 m  when we sample 4096 points from 16384 source points. 
% Sampling subset with average optimum voxel across all scenes, there will be about $20\%$ scenes with less than 3000 non-empty voxels and up to $50\%$ scenes with more than 6000 non-empty voxels, seen in Figure~\ref{fig5-2}. 
% This results in subsets containing redundant points or missing information. 

We recognize the need to select the voxel size for each scene adaptively. 
This strategy helps to retain more of the information present in the raw scene. 
Typically, the smaller the voxel size, the more non-empty voxels in a scene, so more points are sampled. 
Thanks to this monotonically decreasing property, we can determine the most suitable voxel size for a given scene by performing a binary search. 
To this end, we design an adaptive voxel searching (AVS) module to find a suitable voxel size $\mathbf{v}=\textrm{AVS}(\mathcal{P},m)$ for the given point cloud $\mathcal{P}$ and expected sampling number $m$. 
we compute the number of non-empty voxels $m_s$ for a given searched voxel size $\mathbf{v_s}$ by implementing a voxel counter $m_s=\textrm{Cnt}\left(\mathcal{P},\mathbf{v_s}\right)$ in a voxel-hashing manner similar to Section~\ref{sec3-2}. 
During the binary search, the output voxel size will be updated iteratively until the number of non-empty voxels falls into the interval $[m,(1+\sigma)m]$ with tolerance factor $\sigma$, or the iteration times exceeds $t_{max}$. 

\subsection{Hierarchical Paradigm}
So far, we can obtain high-quality sampling results with the above components. 
However, only sampling with a single granularity lacks the ability to capture various scene structures differently. 
For example, some simple geometries like floors and walls only require coarse-grained sampling, whereas pedestrians and cyclists require finer grain. 

Therefore, we introduce a hierarchical point cloud sampling framework that samples the scene at different scales from coarse to fine.
The overall pipeline is illustrated in Figure~\ref{fig2}.
Our framework has k layers, each consisting of an adaptive voxel searching module $\textrm{AVS}$ and center-closest point sampling $\textrm{S}$. 
For layer $\ell$, we obtain $\ell$-th subset by 
\begin{equation} \label{eq:llayersampling}
\begin{aligned}
\mathcal{P}'_\ell=\textrm{S}\left(\mathcal{P}_\ell,\textrm{AVS}\left(\mathcal{P}_\ell,m_\ell\right)\right),
\end{aligned}
\end{equation}
where $\mathcal{P}_\ell$ and $m_\ell$ are the input and sampling number of $\ell$ layer, respectively. 
Then we mask the points in $\mathcal{P}_\ell$ as sampled state with $\mathcal{P}'_\ell$ to form the input of next layer $\mathcal{P}_{\ell+1}$. 
As for $m_\ell$, we set it to be $4 m_{\ell-1}$ and satisfy $m=\sum_{\ell=1}^{k}{m_\ell}$. 
Finally, we gather all subsets $\mathcal{P}'_\ell$ to generate the output $\mathcal{P}'$.
\section{Experiments} 
We conduct extensive experiments for the proposed \modelname{} in this section. 
We first introduce the experimental settings in Section~\ref{sec41}. Then the comparison of runtime shown in Section~\ref{sec42} indicates the efficiency of our method. 
We further evaluate the proposed method on various tasks, including large-scale detection (Section~\ref{sec43}) and segmentation (Section~\ref{sec44}) in both indoor and outdoor scenes, to show the effectiveness.
Finally, we conduct an ablation study in Section~\ref{sec45} to evaluate each component.
% We conduct experiments on ImageNet-1K image classification [18], COCO object detection [39], and ADE20K semantic segmentation [74]. In the following, we first compare the proposed Swin Transformer architecture with the previous state-of-the-arts on the three tasks. Then, we ablate the important design elements of Swin Transformer. 

\subsection{Experimental Settings\label{sec41}} 
We compare outdoor point cloud object detection based on the OpenPCDet~\cite{openpcdet2020} toolbox. For scene segmentation and indoor object detection, we adapt our sampler for the official implementation of each model. All experiments are run on an i7-6700 CPU and a single RTX 2080Ti GPU.

\paragraph{Datasets.} 
The large-scale KITTI~\cite{geiger2012we} and nuscenes~\cite{caesar2020nuscenes} datasets are used for efficiency comparison and outdoor point cloud object detection.  
And SUN RGB-D~\cite{song2015sun} dataset is used to compare point cloud object detection in indoor scenes. 
We conduct outdoor and indoor scene semantic segmentation experiments based on SemanticKITTI~\cite{behley2019semantickitti} dataset and S3DIS~\cite{armeni2017joint} dataset, respectively. 
The splits, setup and pre-processing of each dataset follow previous work.
\paragraph{Implementation Details.}
For all experiments except ablation study, we employ a 2-layer structure for the proposed \modelname{}. For AVS, the tolerance $\sigma$ and max iteration times  $t_{max}$ are set to $0.05$ and 20, respectively. 
We implement parallel voxel-hashing on the GPU via cuHash~\cite{simplecudahash}.

\subsection{Efficiency\label{sec42}}

In this section, we empirically compare the proposed \modelname{} with RPS, RVS and FPS to get an intuitive feel for the efficiency of our method. 

\paragraph{Complexity Analysis}
We first analyze the complexity of these methods and summarise them in Table~\ref{tab:complexity}. 
There is no doubt that RPS has the lowest time and space complexity. 
Conversely, the best FPS has the highest time complexity. 
For ours, sparse voxelization based on voxel-hashing has $\mathcal{O}\left(N\right)$ complexity. 
The center-closest sampling strategy accesses and updates non-empty voxels individually for each point with $\mathcal{O}\left(1\right)$ complexity. 
Similarly, we takes $\mathcal{O}\left(N\right)$ to perform voxel counter $\textrm{Cnt}(\cdot,\cdot)$ for one iteration in AVS. 
While it takes $\mathcal{O}\left(\log{N}\right)$ iterations to find a certain value by binary search, we only need at most 
$\lceil\log{\frac{1}{\sigma}}\rceil$ iterations to converge within a given tolerance $\sigma$. Summarily, our method has a time complexity of $\mathcal{O}(N)$. 

\begin{table}
\begin{center}
\vspace{4pt}
\resizebox{0.8\linewidth}{!}{
\begin{tabular}{l|cccc}
\hline
Method & RPS & RVS & FPS & Ours \\
\hline\hline
Point & 
$\mathcal{O}\left(N\right)$&
$\mathcal{O}\left(N\right)$&
$\mathcal{O}\left(N^2\right)$&
$\mathcal{O}\left(N\right)$\\
Space & 
$\mathcal{O}\left(1\right)$&
$\mathcal{O}\left(N\right)$&
$\mathcal{O}\left(N\right)$&
$\mathcal{O}\left(N\right)$\\
\hline
\end{tabular}
}
\end{center}
\caption{The time and space complexity of different methods. We sample from N points with a typical ratio of 4. 
}\label{tab:complexity}
\end{table}
\begin{table*}[ht]
\begin{center}
\resizebox{0.94\linewidth}{!}{ 
\begin{tabular}{llccc|rrrr|rrrr}
\hline 
\multirow{2}{*}{Dataset} & 
\multirow{2}{*}{Detector}  & &
\multicolumn{2}{c|}{Sampler} & 
\multicolumn{4}{c|}{Performance} &
\multicolumn{4}{c}{Latency (ms) } 
\\
 &
 & &
train & test& 
\multicolumn{2}{c}{Car AP@R11} & \multicolumn{2}{c|}{Car AP@R40} &
\multicolumn{2}{c}{batch size=1} & \multicolumn{2}{c}{batch size=8}
\\
\hline 
\hline 
\multirow{17}{*}{\shortstack{KITTI\\ \textit{val} set}}& 
\multirow{3}{*}{PointRCNN~\cite{shi2019pointrcnn} }
&*& FPS & FPS & 
79.39 &  & 
82.96 &  &
119.9 &  &
75.6 &  \\
&
&& FPS & RPS &
76.89 &\ccr{10}-1.24 &
77.27 &\ccr{20} -3.21 &
94.9 &\ccg{10}-20.9\% &
70.0 &\ccg{5} -7.4\% \\
&
&& FPS & ours &
79.21 &\ccr{5} -0.18 &
82.96 & 0.00 &
95.1 &\ccg{10} -20.7\% &
71.8 &\ccg{5} -5.7\% \\
\colorline{2-13}{gray!50}

&
\multirow{3}{*}{PV-RCNN~\cite{shi2020pv}} 
&*& FPS & FPS &
79.09 &  &
82.68 &  &
117.4 &  &
68.5 & \\
&
&& FPS & RPS &
76.89 &\ccr{15} -2.20 &
77.28 &\ccr{30} -5.37 &
88.0 &\ccg{10} -25.0\% &
47.7 &\ccg{15} -30.4\% \\
&
&& FPS & ours &
79.98 &\ccr{5} -0.11 &
82.53 &\ccr{5} -0.12 &
90.6 &\ccg{10} -22.8\% &
48.4 &\ccg{15} -29.3\% \\
\colorline{2-13}{gray!50}

&
\multirow{3}{*}{3DSSD~\cite{yang20203dssd}}
&*& FPS & FPS &
84.14 &  &
84.54 &  &
84.9 &  &
35.3 &  \\
&
&& FPS & RPS &
78.08 &\ccr{35} -6.06 &
80.03 &\ccr{25} -4.51 &
47.8 &\ccg{20} -36.2\% &
31.7 &\ccg{5} -10.2\% \\
&
&& FPS & ours &
84.26 &\ccg{5} +0.12 &
84.79 &\ccg{5} +0.25 &
49.1 &\ccg{15} -34.5\% &
31.9 &\ccg{5} -9.5\% \\
\colorline{2-13}{gray!50}

&
\multirow{4}{*}{SASA~\cite{chen2022sasa}} 
&*& FPS & FPS &
84.20 &  &
84.67 &  &
63.4 &  &
23.0 &  \\
&
&& FPS & RPS &
77.46 &\ccr{35} -6.73 &
77.76 &\ccr{35} -6.90 &
35.9 &\ccg{20} -43.4\% &
19.8 &\ccg{7} -14.0\% \\
&
&& FPS & ours &
84.54 &\ccg{5} +0.34 &
85.13 &\ccg{5} +0.47 &
39.1 &\ccg{20} -38.3\% &
20.2 &\ccg{6} -12.2\% \\
&
&& ours & ours &
84.42 &\ccg{5} +0.22 &
84.81 &\ccg{5} +0.15 &
39.1 &\ccg{20} -38.3\% &
20.2 &\ccg{6} -12.2\% \\
\colorline{2-13}{gray!50}

&
\multirow{4}{*}{IA-SSD~\cite{zhang2022not}} 
&*& FPS & FPS &
79.22 & &
83.02 & &
46.4 & &
13.8 & \\
&
&& FPS & RPS &
77.29 &\ccr{10} -1.93 &
77.59 &\ccr{30} -5.44 &
19.5 &\ccg{30} -58.0\% &
10.7 &\ccg{10} -22.4\% \\
&
&& FPS & ours &
79.36 &\ccg{5} +0.14 &
83.21 &\ccg{5} +0.19 &
20.1 &\ccg{30} -56.7\% &
10.8 &\ccg{10} -22.0\% \\
&
&& ours & ours &
84.27 &\ccg{30} +5.05 &
85.04 &\ccg{20} +2.02 &
20.1 &\ccg{30} -56.7\% &
10.8 &\ccg{10} -22.0\% \\
\hline
&&&&&
\multicolumn{2}{c}{NDS} & \multicolumn{2}{c|}{mAP} &
\multicolumn{2}{c}{batch size=1} & \multicolumn{2}{c}{batch size=4}\\
\hline
\hline

\multirow{8}{*}{\shortstack{nuScenes\\ \textit{val} set}}&
\multirow{4}{*}{SASA~\cite{chen2022sasa}} 
&*& FPS & FPS &
58.55 &  &
41.24 &  &
500.9 &  &
194.2 &  \\
&
&& FPS & RPS &
54.59 &\ccr{20}  -3.96 &
34.51 &\ccr{35}  -6.73&
164.0 &\ccg{33} -67.3\% &
109.3 &\ccg{20} -43.7\% \\
&
&& FPS & ours &
58.24 &\ccr{5} -0.31 &
40.67 &\ccr{5} -0.57 &
165.1 &\ccg{33} -67.1\% &
109.2 &\ccg{20} -43.8\% \\
&
&& ours & ours &
58.58 &\ccg{5} +0.03 &
41.28 &\ccg{5} +0.04 &
165.1 &\ccg{33} -67.1\% &
109.2 &\ccg{20} -43.8\% \\
\colorline{2-13}{gray!50}

&
\multirow{4}{*}{IA-SSD~\cite{zhang2022not}} 
&*& FPS & FPS &
54.42 & &
37.01 & &
439.0 & &
130.7 & \\
&
&& FPS & RPS &
50.76 &\ccr{20} -3.66 &
31.23 &\ccr{30} -5.66 &
85.1 &\ccg{40} -80.6\% &
45.6 &\ccg{32} -65.1\% \\
&
&& FPS & ours &
54.00 &\ccr{5} -0.42 &
36.22 &\ccr{5} -0.67 &
85.0 &\ccg{40} -80.7\% &
45.7 &\ccg{32} -65.0\% \\
&
&& ours & ours &
55.04 &\ccg{5} +0.62  &
37.01 &\ccg{5} +0.12 &
85.0 &\ccg{40} -80.7\% &
45.7 &\ccg{32} -65.0\% \\
\hline

\end{tabular}
}
\end{center}
\caption{
Performance of outdoor 3D object detection. 
The effect of samplers is evaluated by replacing the first layer sampler of the original models.
We report the 3D AP calculated by both 11 and 40 recall positions for KITTI.
For nuScenes, we follow the standard protocol~\cite{caesar2020nuscenes} by reporting NDS and mAP.  
Besides, the latency measured with test with different batch size is also shown.
*: the original configuration.
}
\label{tab:kitti}
\end{table*}

\paragraph{Latency Comparison}
\begin{figure}[!b]
\centering
\includegraphics[width=1.0\linewidth]{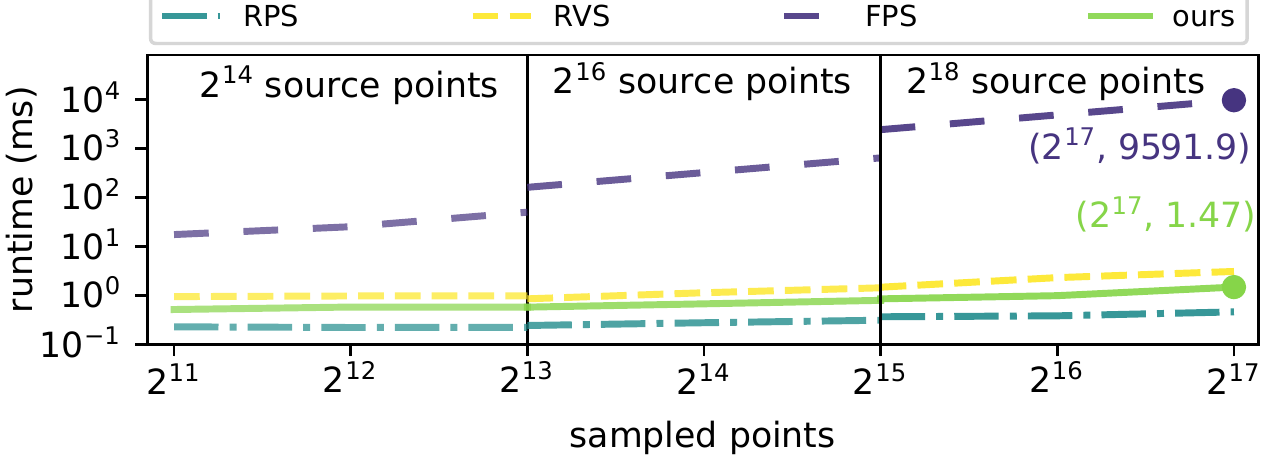}
\caption{
We conduct 3 groups of experiments with different scale inputs, (left) small scale with $2^{14}$ points, (mid) medium scale with $2^{16}$ points and (right) $2^{18}$ points for large scale. 
In each group, the point cloud is downsampled by 3 steps with a ratio of 2 at each step. 
Green solid line indicates ours.
}
\label{fig7}
\end{figure}
Figure~\ref{fig7} shows the results of our latency comparison. 
We conduct experiments on 3 different input scales to compare the runtimes in different application scenarios: 
\begin{enumerate*}[label=(\arabic*)]
\item small-scale scenarios with $2^{14}$ points such as the KITTI dataset;
\item $2^{16}$-point medium-sized scenes like the nuSCenes dataset;
\item large-scale $2^{18}$ points, the same order of magnitude as the output of 64-beam LiDAR. 
\end{enumerate*}
And all the methods run in parallel on the GPU. 

Unsurprisingly, RPS is the fastest method, while the heavy FPS has the highest latency due to the quadratic complexity and difficulty of parallelism. 
FPS takes more than 9 seconds to sample from $2^{18}$ points to $2^{17}$ points.
This means that it cannot be used to directly process the original point cloud output from LiDAR.
It can be noticed that for all experimental configurations, our method can run within 1.47ms. 
Moreover, our \modelname{} is even fast than RVS, which is also based on the voxels. 
Experimental results show that our algorithm has efficiency on the same level as RPS and RVS, and is $10^2$ to $10^4$ times faster than FPS.

\subsection{Point Cloud Object Detection\label{sec43}}
In this section, We evaluate the impact of the proposed \modelname{} on 3D object detection task. For a fair comparison, we test all detectors on the same platform. 

\begin{table*}[ht]
\begin{center}
\resizebox{1.0\linewidth}{!}{ 
\begin{tabular}{lccc|c|rrrrrrrrrr}
\hline 
\multirow{2}{*}{Detector} && \multicolumn{2}{c|}{Sampler} &  
\multirow{2}{*}{mAP@0.5} &
\multirow{2}{*}{bed} & 
\multirow{2}{*}{tables} & 
\multirow{2}{*}{sofa} &
\multirow{2}{*}{chair} & 
\multirow{2}{*}{toilet} & 
\multirow{2}{*}{desk} &
\multirow{2}{*}{dresser} & 
\multirow{2}{*}{nightstand} & 
\multirow{2}{*}{bookshelf} &
\multirow{2}{*}{bathtub}\\ 
&&train&test&&&&&&&&&&&
\\
\hline
VoteNet~\cite{qi2019deep}&
*&
FPS&FPS&
33.32	&
47.98	&
19.44	&
42.35	&
53.78	&
63.97	&
5.29    &
15.7	&
35.86	&
4.54    &
45.24	\\
&
&
FPS&ours&
32.82	&
46.87	&
20.25	&
42.93	&
53.17	&
56.84	&
5.12	&
14.93	&
37.18	&
4.89    &
46.04 \\
&
&
\multicolumn{2}{c|}{\textit{improve}}&
\ccr{5}-0.50   &
\ccr{11}-1.12   &
\ccg{8}+0.81   &
\ccg{5}+0.58   &
\ccr{6}-0.61   &
\ccr{40}-6.12   &
\ccr{2}-0.17   &
\ccr{8}-0.80   &
\ccg{13}+1.32   &
\ccg{4}+0.35   &
\ccg{8}+0.81  \\
\hline
3DETR~\cite{misra2021end}
&
*&
FPS&FPS&
30.73	&
48.13	&
18.23	&
40.74	&
44.65	&
66.89	&
7.93    &
12.33	&
31.13	&
4.96    &
32.32   \\
&
&
FPS&ours&
32.01	&
51.50	&
20.00	&
39.17	&
45.61	&
64.48	&
5.90	&
14.67	&
34.52	&
5.66    &
38.88   \\
&
&
\multicolumn{2}{c|}{\textit{improve}}&
\ccg{13}+1.28 &
\ccg{30}+3.37   &
\ccg{17}+1.77   &
\ccr{16}-1.57   &
\ccg{10}+0.96   &
\ccr{24}-2.41   &
\ccr{20}-2.03   &
\ccg{23}+2.34   &
\ccg{33}+3.39   &
\ccg{7}+0.70   &
\ccg{40}+6.56  \\
\hline
\end{tabular}
}
\end{center}
\caption{
Performance of indoor 3D object detection on SUN RGN-D dataset. 
We replace the first layer sampler of the original models. 
The reported mAP is calculated at the more strict IoU threshold of 0.5. 
*: the original configuration.
}
\label{tab:sunrgbd}
\end{table*}
\paragraph{Outdoor Scenes} We first evaluate our \modelname{} on KITTI \textit{val} set and report the results in Table~\ref{tab:kitti}. 
To compare the effect of samplers on detector performance, we replace the first layer of samplers in the original model with different sampling methods. Comparing the first and second rows in Table~\ref{tab:kitti}, When we directly replace the FPS with RPS in models and test without retraining, the 3D AP at both R11 and R40 drops significantly for all models. 
Especially, SASA has a performance degradation with 6.73 AP and 6.90 AP at R11 and R40, respectively. 
However, we observe that the detection performance is unchanged or even slightly higher than FPS when tested directly test with our \modelname{}. 
It should be noted that the follow-up networks of samplers are sensitive to the distribution of sampled points after training, so performance will commonly degrade when tested with different samplers. 
The similar performance between our method and FPS means that the proposed \modelname{} has a distribution close to that of FPS, which verifies our observations in Section~\ref{sec3-1}. 
In addition, our method can greatly reduce detector runtimes by up to 56.7\%, meaning that our method is highly efficient. 
We also retrain two lightweight models, SASA and IA-SSD, with our \modelname{}. 
Comparing the first and last rows of them, the performance improved after retraining. 
Most surprisingly, our approach makes the IA-SSD improve by a large margin of 5.05\% AP@R11 and can infer at a speed of 50 Hz. 

We also experiment on larger scenes dataset, nuScenes, and report the results at the bottom of Table~\ref{tab:kitti}. 
With similar results on KITTI dataset, RPS has the worst performance. 
Our method yields a slight performance gain after retraining. 
Notably, the efficiency advantage of our method is even more remarkable when dealing with larger point clouds, saving 80.7\% of time consumption.

\begin{figure*}[!t]
\begin{center}
    
\subfigure{ 
\setcounter{subfigure}{0}
\begin{minipage}[]{0.01\linewidth}
    \rotatebox{90}{
        \footnotesize{
            \quad
            Car \quad \quad \quad  \quad \quad  \quad \quad  
            Cyclist \quad \quad \quad  \quad  \quad
            Pedestrian \quad 
        }
    }
\end{minipage}
}
\subfigure[source] { 
\begin{minipage}[]{0.17\linewidth}
    \includegraphics[width=1.0\linewidth]{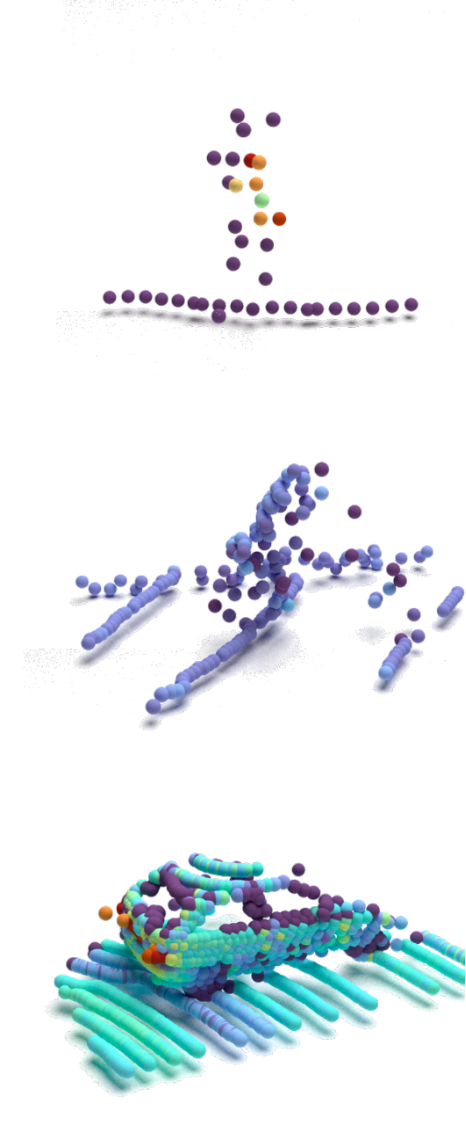}
\end{minipage}
}
\hspace{3pt}
\subfigure[RPS] { 
\begin{minipage}[]{0.17\linewidth}
    \includegraphics[width=1.0\linewidth]{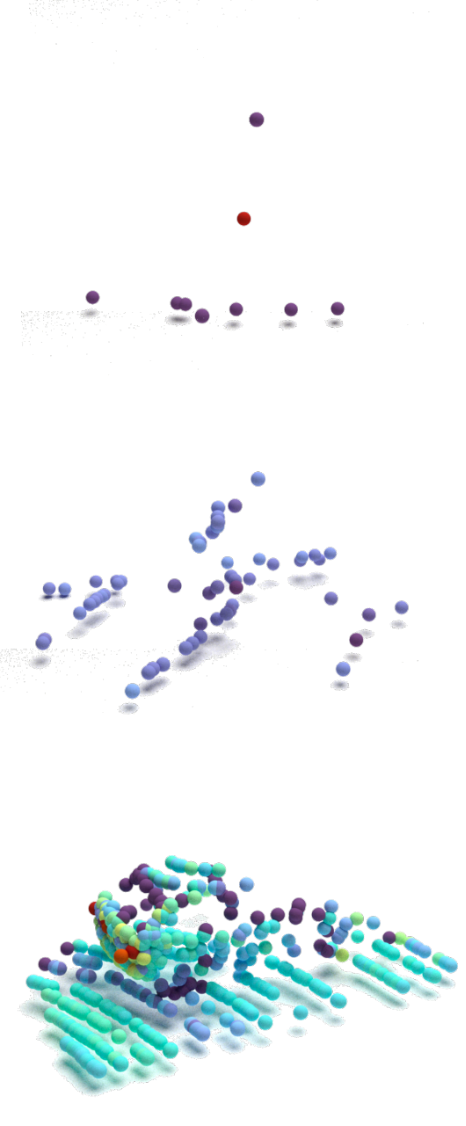}
\end{minipage}
}
\hspace{3pt}
\subfigure[RVS] { 
\begin{minipage}[]{0.17\linewidth}
    \includegraphics[width=1.0\linewidth]{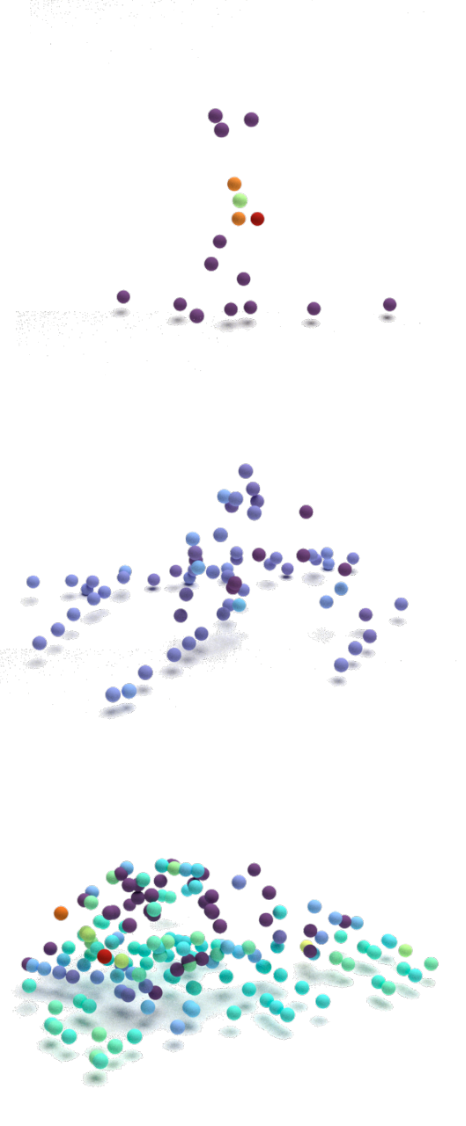}
\end{minipage}
}
\hspace{3pt}
\subfigure[FPS] { 
\begin{minipage}[]{0.17\linewidth}
    \includegraphics[width=1.0\linewidth]{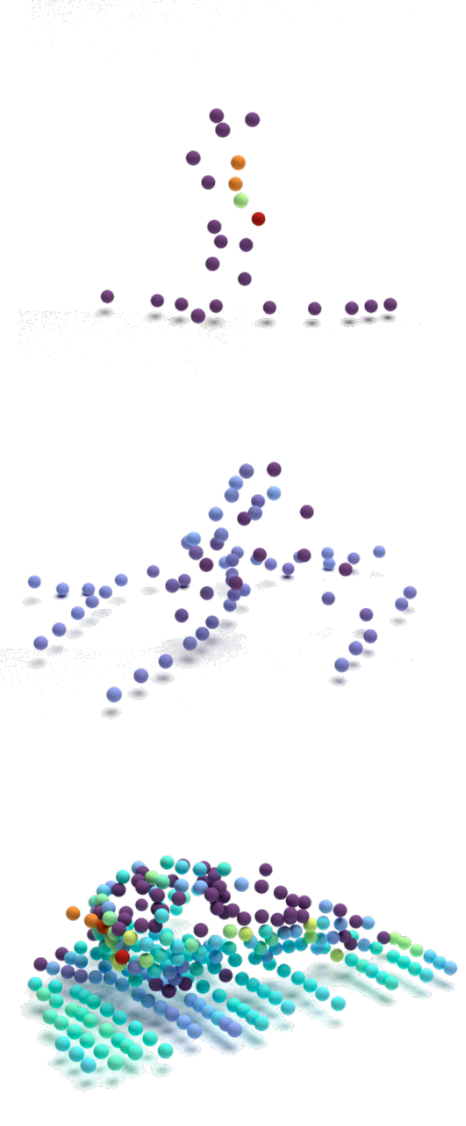}
\end{minipage}
}
\hspace{3pt}
\subfigure[Ours] { 
\begin{minipage}[]{0.17\linewidth}
    \includegraphics[width=1.0\linewidth]{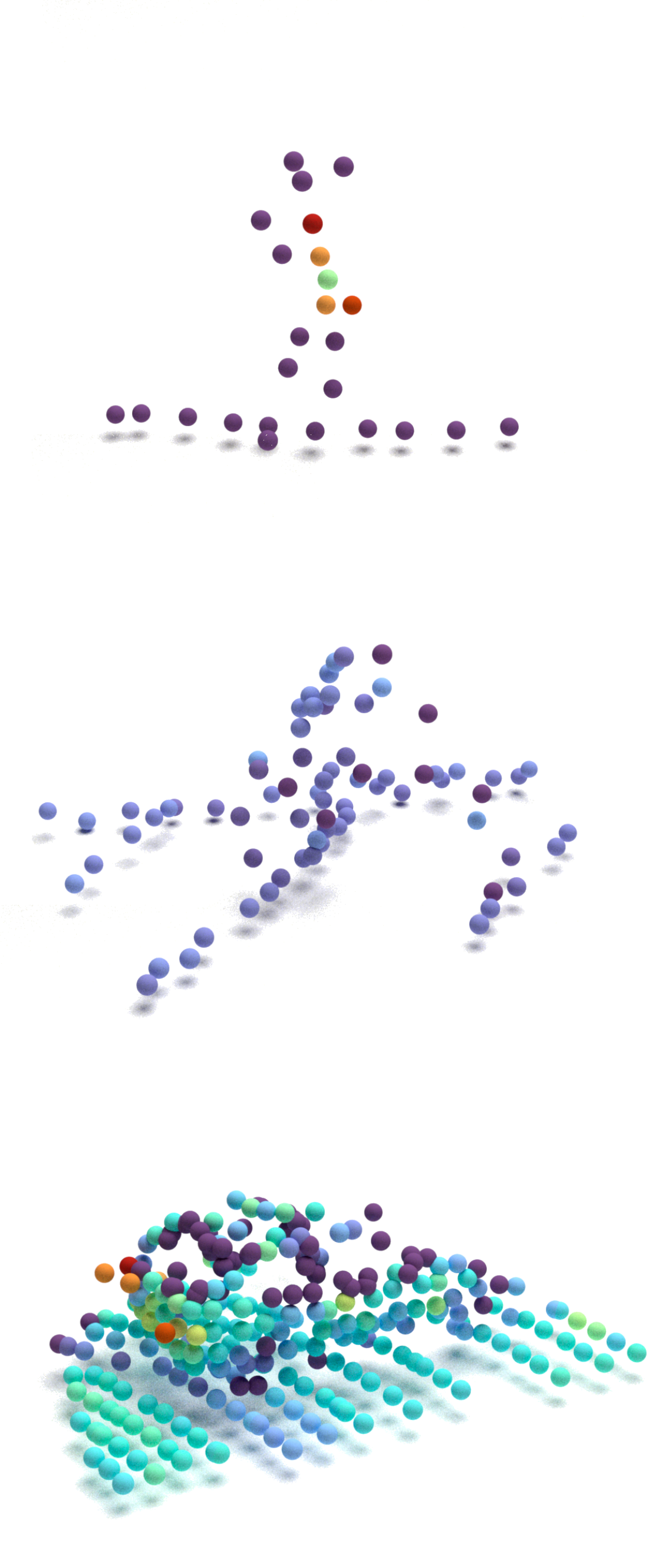}
\end{minipage}
}

\end{center}
\caption{
Examples of different sampling results. From top to bottom: a near car, a medium distance cyclist, a distant pedestrian. 
}
\label{fig:vis}
\end{figure*}

In summary, our method can be adapted to large-scale outdoor 3D object detection tasks. 
As well as saving time, HAVSampler has the potential to boost performance. 
\vspace{-2.5pt}
\paragraph{Indoor Scenes}
Instead of LiDAR, indoor data sets are captured by RGB-D cameras. 
We evaluated our method on the SUN RGB-D dataset to verify its applicability to different sensor data and show results in Table~\ref{tab:sunrgbd}. 
We report the more strict mAP calculated at IoU threshold of 0.5. 
In VoteNet~\cite{qi2019deep}, our sampling strategy has a performance degradation of 0.5 mAP. 
However, an increase of 1.28 mAP is obtained for transformer-based 3D detector 3DETR~\cite{misra2021end}. 
This subsection manifests again that our samplers have a similar performance with FPS on detection tasks. 
\vspace{-3pt}
\paragraph{Recall Analysis}
\begin{table}[b]
\vspace{-10pt}
\begin{center}
\resizebox{0.9\linewidth}{!}{
\begin{tabular}{r|cccc}
\hline
Method & RPS & RVS & FPS & Ours \\
\hline\hline
Point recall & 
24.99\%&
19.87\%&
18.16\%&
18.56\%\\
Instance recall & 
98.18\%&
99.90\%&
99.94\%&
99.96\%\\
\hline
\end{tabular}
}
\end{center}
\caption{Recall analysis on KITTI dataset. 
}\label{tab:pointrecall}
\end{table}
We downsampled the points to 1/4 of the input and analyzed the recall of different samplers on the KITTI dataset. 
Point recall is the ratio of points belonging to foreground objects existed in the subset. 
Instance recall refers to the percentage of instances that contain more than 1 point in the subset. 
Although RPS retains the most foreground points, it has the lowest instance recall due to the loss of distant points, which can be visualized in  Figure~\ref{fig:vis}. 
Compared to FPS, our method has a higher point recall while still keeping more instances after downsampling, which is important for detection tasks.

\subsection{Point Cloud Scene Segmentation\label{sec44}}
\begin{table}
\begin{center}
\resizebox{1.0\linewidth}{!}{ 
\begin{tabular}{c|clccc|rr}
\hline 
&
\multirow{2}{*}{\small{Dataset}} & 
\multirow{2}{*}{\small{Model}}  & 
&
\multicolumn{2}{c|}{\small{Sampler}} & 
\multicolumn{2}{c}{\multirow{2}{*}{mIoU}} \\
& &   &   &   train   &   test    &   &   \\
\hline 
\hline 
\multirow{9}{*}{\rotatebox{90}{\small{\shortstack{Outdoor}}}}&
\multirow{9}{*}{\rotatebox{90}{\small{\shortstack{\\SemanticKITTI~\cite{behley2019semantickitti} \\ sequence 8}}}}
&\small{RandLA-Net~\cite{hu2020randla}}      &*& RPS  & RPS  & 50.0 &        \\
&&                                           & & RPS  & ours & 51.0 & \ccg{25}+1.0   \\
&&                                           & & ours & ours & 54.1 & \ccg{41}+4.1   \\
\colorline{3-8}{gray!50}
&
&\small{BAAF-Net~\cite{qiu2021semantic}}     &*& FPS  & FPS  & 48.6 &        \\
&&                                           & & FPS  & ours & 49.2 & \ccg{20}+0.6   \\
&&                                           & & ours & ours & 51.7 & \ccg{31}+3.1   \\
\colorline{3-8}{gray!50}
&
&\small{BAF-LAC~\cite{shuai2021backward}}    &*& RPS  & RPS  & 47.8 &        \\
&&                                           & & RPS  & ours & 47.6 & \ccr{10}-0.2   \\
&&                                           & & ours & ours & 48.0 & \ccg{10}+0.2   \\
\hline
\multirow{6}{*}{\rotatebox{90}{\small{\shortstack{Indoor}}}}&
\multirow{6}{*}{\rotatebox{90}{\small{\shortstack{S3DIS~\cite{armeni2017joint}\\6-fold CV}}}}
&\small{RandLA-Net~\cite{hu2020randla}}     &*& RPS  & RPS  & 70.1 &        \\
&&                                           & & RPS  & ours & 70.3 & \ccg{15}+0.3   \\
\colorline{3-8}{gray!50}
&&\small{BAAF-Net~\cite{qiu2021semantic}}    &*& FPS  & FPS  & 72.1 &        \\
&&                                           & & FPS  & ours & 72.0 & \ccr{5}-0.1   \\
\colorline{3-8}{gray!50}
&&\small{SCF-Net~\cite{fan2021scf}}          &*& RPS  & RPS  & 71.6 &        \\
&&                                           & & RPS  & ours & 71.7 & \ccg{5}+0.1   \\
\hline
\end{tabular}
}
\end{center}
\caption{
Performance of both outdoor and indoor point cloud scene segmentation. 
We report the mean IoU on SemanticKITTI \textit{val} set. 
As for S3DIS, we compute the mean IoU by the 6-fold cross validation.
*: the original configuration.
}
\label{tab:semkitti}
\end{table}
The scene segmentation task densely predicts per-point semantics for large-scale point clouds, requiring greater efficiency and sampling quality than detection.
\vspace{-5pt}
\paragraph{Outdoor Scenes}
The results at the top of table~\ref{tab:semkitti} show the advantages of our method in scene segmentation. 
Our sampling strategy outperforms both RPS and FPS by a large margin, improving by 4.1\% and 3.1\% mIoU after retraining, respectively. 
For BAF-LAC, however, there is no noticeable change compared to RPS. 
We guess it is mainly because the backward attention fusing mechanism in BAF-LAC~\cite{shuai2021backward} complements the points lost by RPS with skip connections.

\vspace{-5pt}
\paragraph{Indoor Scenes}
The rest of Table~\ref{tab:semkitti} shows the indoor segmentation results. 
I find that our samplers have less performance improvement indoors. 
In all models, it brings no more than 0.3\% mIoU change. 
We suspect that the main reason for this is that RGB-D captures denser point clouds indoors than outdoor LiDAR. 
So there has not been much change in either the replacement of FPS or RPS. 
\vspace{0pt}

\subsection{Ablation Study\label{sec45}}
\begin{table}
\begin{center}
\resizebox{1.0\linewidth}{!}{ 
\begin{tabular}{c|c|cccc|cc|cc}
\hline 
\multirow{2}{*}{\#} &
\multirow{2}{*}{L} &
\multicolumn{4}{c|}{Point section strategy} &  
\multicolumn{2}{c|}{Voxel} &  
\multicolumn{2}{c}{Moderate Car} \\
&    & ME           &           RA &           CE &           CC &Fix           &Ada           & AP\small{@R11}    & AP\small{@R40} \\
\hline
\hline
\multirow{2}{*}{\textcolor{red!50}1}
&\ccr{20}1
     & \ccr{20}$\checkmark$ 
                    &    \ccr{20}  &    \ccr{20}  &    \ccr{20}  & \ccr{20}$\checkmark$ 
                                                                                &\ccr{20}      &\ccr{20}79.29
                                                                                                           & \ccr{20}83.04 \\
&1   & $\checkmark$ &              &              &              &              & $\checkmark$ & 83.81     & 84.49 \\
\colorline{1-10}{gray!30}

\multirow{2}{*}{\textcolor{red!50}2}
&1   &              & $\checkmark$ &              &              & $\checkmark$ &              & 79.03     & 82.69 \\
&1   &              & $\checkmark$ &              &              &              & $\checkmark$ & 79.45     & 83.24 \\
\colorline{1-10}{gray!30}

\multirow{2}{*}{\textcolor{red!50}3}
&1   &              &              & $\checkmark$ &              & $\checkmark$ &              & 79.27     & 82.99 \\
&1   &              &              & $\checkmark$ &              &              & $\checkmark$ & 83.81     & 84.49 \\
\colorline{1-10}{gray!30}

\multirow{2}{*}{\textcolor{red!50}4}
&1   &              &              &              & $\checkmark$ & $\checkmark$ &              & 79.24     & 82.93 \\
&1   &              &              &              & $\checkmark$ &              & $\checkmark$ & 84.08     & 84.67 \\
\hline

\multirow{2}{*}{\textcolor{red!50}5}
&\ccb{20}2   
      &  \ccb{20}   &  \ccb{20}     &   \ccb{20}  &\ccb{20}$\checkmark$
                                                                  &\ccb{20}     & \ccb{20}$\checkmark$ 
                                                                                               &\ccb{20}84.27
                                                                                                           & \ccb{20}85.04 \\
&3   &              &              &              & $\checkmark$ &              & $\checkmark$ & 79.20     &         83.04 \\
\hline
\end{tabular}
}
\end{center}
\caption{
Effects of different strategies of layer, point selection and voxel initialisation for \modelname{}. 
ME: mean. 
RA: random. 
CE: centre. 
CC: center-closest. 
Fix: Fixed voxel size.
Ada: Adaptive voxel size. 
The red row is baseline RVS. 
The row in blue is the best configuration for the proposed method. 
}
\label{tab:ablation}
\end{table}
We use IA-SSD detector and KITTI dataset for quick experiments. 
The results of car detection are shown in Table~\ref{tab:ablation}. 
From groups 1-4, we can see that the random selection of a representative point in voxels has the worst performance. 
Besides, when using adaptive voxels, our most center-closest point selection helps the detector to obtain higher APs.
For all point selection strategies, adaptive voxels bring major performance improvements. 
Group 5 investigates how the number of layers in our architecture affects performance. 
The experimental results indicate that a 2-layer structure can achieve the best performance. 
\subsection{Qualitative Results and Discussion\label{sec4-6}}
Figure~\ref{fig:vis} shows examples of point clouds subset from different samplers. 
In line with Table~\ref{tab:pointrecall}, RPS retains most of the points close to the sensor, but loses most points further away. 
In contrast, RVS can prevent the distant points from being abandoned but has messy sampling results at the near place. 
From the last two columns, it can be observed that both FPS and our \modelname{} well deal with the point cloud at all distances and preserve the skeleton of source point clouds when downsampling.

\section{Limitations}
Although the performance of efficient \modelname{} is competitive with or even higher than the state-of-the-art FPS on object detection and scene segmentation. 
We find that compared to indoor RGB-D point clouds, \modelname{} performs better for outdoor LiDAR point clouds. 
In addition, our experiments on point cloud classification and partition segmentation show that there is a counter-intuitive drop in performance when replacing the FPS with ours, which can be seen in the supplementary material. 
But this performance degradation is still much smaller than using RPS. 
However, the sampler latency in these object-level tasks is ignorable and so is not a concern for us\looseness=-1. 

\section{Conclusion}
% 我们分析了分析了...的优点和缺点。不同于以往点云密度的角度，发现均匀的间距是高性能的关键。其可以保留原始场景的几何结构，这对下游任务如检测和分割的性能关键。基于此观察我们提出利用网格天然间距均匀的特点，对距离网格中心最近的那个点来进行采样。接着我们通过分层采样和根据场景大小自适应的选取何时的初始化体素大小，来充分对原始场景进行采样。
% 最终，我们多种任务下的实验表明我们的算法在比最先进的FPS采样效率高两个数量级，且部分结果超越了它。
% 我们的方法以线性复杂度且高可并行的方式取得了比拟FPS的性能。尤其是在xx任务上取得了...的效率且超过了FPS的表现。
% 我们的算法突破了点云法因采样模块在大规模点云中的效率问题而无法被应用的瓶颈，特别是对于资源受限和实时应用而言。
% 我们希望我们的研究可以让社区对点云法重振信心，设计出更好的点云学习网络。 We believe

We present \modelname{}, an highly efficient yet powerful sampler for real-time applications in large-scale point clouds. 
We reveal that evenly spacing the points in the subset is crucial for optimizing the performance of downstream tasks 
Based on this conclusion, our design of grid-guided paradigm, adaptive voxelization, and hierarchical architecture results in outstanding performance while reducing inference time by 20-80\%. 
This breakthrough in efficiency addresses the bottleneck of the sampling step in real-time applications. 
We release our source code and hope that it can serve as a solid component to inspire and encourage the point cloud learning community. 
\newpage

{\small
\bibliographystyle{ieee_fullname}
\bibliography{egbib}
}
 
\end{document}